\documentclass{article}
\usepackage{spconf}
\ninept

\usepackage[utf8]{inputenc} 
\usepackage[T1]{fontenc}    
\usepackage{hyperref}       
\usepackage{url}            
\usepackage{booktabs}       
\usepackage{amsfonts}       
\usepackage{nicefrac}       
\usepackage{microtype}      
\usepackage{siunitx}

\hypersetup{allcolors=blue,colorlinks=true}

\usepackage{graphicx}
\usepackage{amsmath}
\usepackage{amsthm}
\usepackage{amssymb}
\usepackage{amsfonts}
\usepackage{mathrsfs}
\usepackage{multirow}

\usepackage[utf8]{inputenc}
\usepackage[natbib=true,
            bibstyle=ieee,
            citestyle=ieee,
            uniquename=false,
            uniquelist=false,
            maxcitenames=2,
            mincitenames=1]{biblatex}
\renewbibmacro{in:}{\ifentrytype{article}{}{\printtext{\bibstring{in}\intitlepunct}}}
\usepackage{xspace}

\newcommand{\clustRviz}{{\tt clustRviz}\xspace}
\newcommand{\cobra}{{\tt COBRA}\xspace}

\usepackage{bm}

\newcommand{\bX}{\bm{X}}
\newcommand{\bU}{\bm{U}}
\newcommand{\bV}{\bm{V}}
\newcommand{\bZ}{\bm{Z}}

\newcommand{\bC}{\bm{C}}
\newcommand{\bD}{\bm{D}}

\newcommand{\bI}{\bm{I}}
\newcommand{\bw}{\bm{w}}
\newcommand{\bW}{\bm{W}}
\newcommand{\bA}{\bm{A}}
\newcommand{\bB}{\bm{B}}
\newcommand{\bb}{\bm{b}}

\newcommand{\bu}{\bm{u}}

\newcommand{\bv}{\bm{v}}

\newcommand{\by}{\bm{y}}
\newcommand{\bz}{\bm{z}}
\newcommand{\bzero}{\bm{0}}

\usepackage{bbm}

\newcommand{\R}{\mathbb{R}}

\DeclareMathOperator{\prox}{\textsf{prox}}
\DeclareMathOperator{\proj}{\textsf{proj}}

\DeclareMathOperator*{\argmin}{arg\,min}

\DeclareMathOperator*{\Tr}{Tr}

\renewcommand{\(}{\begin{equation}}
\renewcommand{\)}{\end{equation}}

\usepackage{enumerate}
\usepackage{algorithm}
\usepackage{setspace}

\theoremstyle{remark}

\DefineBibliographyStrings{english}{url = Available at}

\title{Splitting Methods for Convex Bi-Clustering and Co-Clustering}

\name{Michael Weylandt\thanks{MW acknowledges support from the NSF Graduate Research Fellowship Program under grant number 1450681.}}

\address{%
  Department of Statistics \\
  Rice University \\
  Houston, TX 77005 \\
  \href{mailto:michael.weylandt@rice.edu}{\tt michael.weylandt@rice.edu}
}

\begin{document}
\maketitle
\begin{abstract}
  Co-Clustering, the problem of simultaneously identifying clusters across multiple aspects of a data set, is a natural generalization of clustering to higher-order structured data. Recent convex formulations of bi-clustering and tensor co-clustering, which shrink estimated centroids together using a convex fusion penalty, allow for global optimality guarantees and precise theoretical analysis, but their computational properties have been less well studied. In this note, we present three efficient operator-splitting methods for the convex co-clustering problem: a standard two-block ADMM, a Generalized ADMM which avoids an expensive tensor Sylvester equation in the primal update, and a three-block ADMM based on the operator splitting scheme of Davis and Yin. Theoretical complexity analysis suggests, and experimental evidence confirms, that the Generalized ADMM is far more efficient for large problems. \\
\end{abstract}

\section{Introduction}

\subsection{Convex Clustering, Bi-Clustering, and Co-Clustering}

In recent years, there as been a surge of interest in the expression of classical statistical problems as penalized estimation problems. These convex re-formulations have many noteworthy advantages over classical methods, including improved statistical performance, analytical tractability, and efficient and scalable computation. In particular, a convex formulation of clustering \citep{Pelckmans:2005,Hocking:2011,Lindsten:2011} has sparked much recent research, as it provides theoretical and computational guarantees not available for classical clustering methods \citep{Zhu:2014,Tan:2015,Radchenko:2017}. Convex clustering combines a squared Frobenius norm loss term, which encourages the estimated centroids to remain near the original data, with a convex fusion penalty, typically the $\ell_q$-norm of the row-wise differences, which shrinks the estimated centroids together, inducing clustering behavior in the solution: 
\begin{equation}
    \hat{\bU} = \argmin_{\bU \in \R^{n \times p}} \frac{1}{2}\|\bX - \bU\|_F^2 + \lambda \sum_{\substack{i, j = 1 \\ i < j}}^n w_{ij} \|\bU_{i\cdot} - \bU_{j\cdot}\|_q \label{eqn:clust}
\end{equation}
Observations $i$ and $j$ are assigned to the same cluster if $\hat{\bU}_{i\cdot} = \hat{\bU}_{j\cdot}$. 

For highly structured data, it is often useful to cluster both the rows and the columns of the data. For example, in genomics one may wish to simultaneously estimate disease subtypes (row-wise clusters of patients) while simultaneously identifying genetic pathways (column-wise clusters of genes). While these clusterings can be performed independently, it is often more efficient to jointly estimate the two clusterings, a problem known as \emph{bi-clustering}. Building on \eqref{eqn:clust}, \citet{Chi:2017} propose a convex formulation of bi-clustering, given by the following optimization problem: 
\begin{equation}
\begin{aligned}
\hat{\bU} &= \argmin_{\bU \in \R^{n \times p}} \frac{1}{2}\|\bX - \bU\|_F^2 \\ & +\lambda \left(\sum_{\substack{i, j = 1 \\ i < j}}^n w_{ij} \|\bU_{i\cdot} - \bU_{j\cdot}\|_q \right.  \left.+ \sum_{\substack{k, l = 1 \\ k < l}}^p \tilde{w}_{kl}\|\bU_{\cdot k} - \bU_{\cdot l}\|_q\right)
\end{aligned} \label{eqn:bclust}
\end{equation}
for fixed $\lambda \in \R_{\geq 0}$, $\bw \in \R^{\binom{n}{2}}_{\geq 0}$, and $\tilde{\bw} \in \R^{\binom{p}{2}}_{\geq 0}$. As with standard convex clustering, the fusion penalties fuse elements of $\bU$ together, but here the rows and columns are both fused together, resulting in a characteristic checkerboard pattern in $\hat{\bU}$.

Extending this, \citet{Chi:2018b} propose a convex formulation of \emph{co-clustering}, the problem of jointly clustering along each mode of a tensor (data array). Their estimator is defined by the following optimization problem, where both $\mathcal{X}$ and $\mathcal{U}$ are order $J$ tensors of dimension $n_1 \times n_2 \times \cdots \times n_J$ and $\mathcal{X}^{j/i}$ denotes the $i$\textsuperscript{th} slice of $\mathcal{X}$ along the $j$\textsuperscript{th} mode:
\begin{equation}
  \hat{\mathcal{U}} = \argmin_{\mathcal{U}} \frac{1}{2}\|\mathcal{X} - \mathcal{U}\|_F^2 + \lambda \sum_{j=1}^J \sum_{\substack{k, l = 1 \\ k < l}}^{n_J}w^j_{k, l}\|\mathcal{U}^{j/k} - \mathcal{U}^{j/l}\|_q \label{eqn:cclust}
\end{equation}
for fixed $\lambda \in \R_{\geq 0}$ and $\bw^j \in \R^{\binom{n_j}{2}}_{\geq 0}, j = 1, \dots, J$, and where $\|\mathcal{X}\|_q$ denotes the $\ell_q$-norm of the vectorization of $\mathcal{X}$. While many algorithms have been proposed to solve Problem \eqref{eqn:clust}, there has been relatively little work on efficient algorithms to solve Problems \eqref{eqn:bclust} and \eqref{eqn:cclust}. Operator splitting methods have been shown to be among the most efficient algorithms for convex clustering \citep{Chi:2015} and, as we will show below, they are highly-efficient for bi-clustering and co-clustering as well.

\subsection{Operator Splitting Methods}

Many problems in statistical learning, compressive sensing, and sparse coding can be cast in a ``loss + penalty'' form, where a typically smooth loss function, measuring the in-sample accuracy, is combined with a structured penalty function (regularizer) which induces a simplified structure in the resulting estimate to improve performance. The popularity of this two-term structure has spurred a resurgence of interest in so-called operator splitting methods, which break the problem into simpler subproblems which can be solved efficiently \citep{Glowinski:2016}. Among the most popular of these methods is the Alternating Direction Method of Multipliers (ADMM), which is typically quite simple to implement, has attractive convergence properties, is easily parallelizable and distributed, and has been extended to stochastic and accelerated variants \citep{Boyd:2011,Ouyang:2013,Goldstein:2014}. 

The ADMM can be used to solve problems of the form
\(\argmin_{(\bu, \bv) \in \mathcal{H}_1 \times \mathcal{H}_2} f(\bu) + g(\bv) \quad \text{ subject to } \quad \mathfrak{L}_1\bu + \mathfrak{L}_2\bv = \bb \label{eqn:two_block}\) where $\bu$, $\bv$ take values in arbitrary (real, finite-dimensional) Hilbert spaces $\mathcal{H}_1$, $\mathcal{H}_2$, $\mathfrak{L}_i: \mathcal{H}_i \to \mathcal{H}_*, i = 1, 2$ are linear operators from $\mathcal{H}_i$ to a common range space $\mathcal{H}_*$, $\bb$ is a fixed element of $\mathcal{H}_*$, and $f, g$ are closed, proper, and convex functions from $\mathcal{H}_1$ and $\mathcal{H}_2$ to $\overline{\R} = \R \cup\{\infty\}$. The ADMM proceeds by Gauss-Seidel updates of the augmented Lagrangian 
\begin{align*}
  \mathscr{L}_{\rho}(\bu, \bv, \bz) &= f(\bu) + g(\bv) + \langle \bz, \mathfrak{L}_1\bu + \mathfrak{L}_2\bv - \bb\rangle_{\mathcal{H}_*} \\ &\quad + \frac{\rho}{2}\|\mathfrak{L}_1\bu + \mathfrak{L}_2\bv - \bb\|_{\mathcal{H}_*}^2.
\end{align*} The ADMM steps are given by: 
\begin{subequations}
\renewcommand{\theequation}{\theparentequation.\arabic{equation}}
\begin{align}
  \bu^{(k+1)} &= \argmin_{\bu \in \mathcal{H}_1} \mathscr{L}_{\rho}(\bu, \bv^{(k)}, \bz^{(k)}) \label{eqn:admm_1}\\
  \bv^{(k+1)} &= \argmin_{\bv \in \mathcal{H}_2} \mathscr{L}_{\rho}(\bu^{(k+1)}, \bv, \bz^{(k)})  \label{eqn:admm_2} \\
  \bz^{(k+1)} &= \bz^{(k)} + \rho(\mathfrak{L}_1\bu^{(k+1)} + \mathfrak{L}_2\bv^{(k+1)} - \bb).
\end{align}
\end{subequations}
While this level of generality is not typically required in applications, we will see below that judicious use of the general formulation is key to developing efficient ADMMs for the co-clustering problem.

While the simplified sub-problems are often much easier to solve than the original problem, they may still be impractically expensive. \citet{Deng:2016} consider a \emph{generalized} ADMM, which augments the subproblems (\ref{eqn:admm_1}-\ref{eqn:admm_2}) with positive-definite quadratic operators $\mathfrak{A}$ and $\mathfrak{B}$ to obtain the updates: 
\begin{subequations}
\renewcommand{\theequation}{\theparentequation.\arabic{equation}}
\begin{align}
  \bu^{(k+1)} &= \argmin_{\bu \in \mathcal{H}_1} \mathscr{L}_{\rho}(\bu, \bv^{(k)}, \bz^{(k)}) + \mathfrak{A}(\bu - \bu^{(k)})\label{eqn:gadmm_1}\\
  \bv^{(k+1)} &= \argmin_{\bv \in \mathcal{H}_2} \mathscr{L}_{\rho}(\bu^{(k+1)}, \bv, \bz^{(k)}) + \mathfrak{B}(\bv - \bv^{(k)})\label{eqn:gadmm_2} \\
  \bz^{(k+1)} &= \bz^{(k)} + \rho(\mathfrak{L}_1\bu^{(k+1)} + \mathfrak{L}_2\by^{(k+1)} - \bb).
\end{align}
\end{subequations}
If $\mathfrak{A}$ and $\mathfrak{B}$ are appropriately chosen, the generalized subproblems (\ref{eqn:gadmm_1}-\ref{eqn:gadmm_2}) may be easier to solve than their standard counterparts.

The wide adoption of the ADMM has lead to many attempts to extend the ADMM to objectives with three or more terms. 
Recently, \citet[Algorithm 8]{Davis:2017} proposed an operator-splitting method for the three-block problem:
\begin{equation}
\begin{aligned}
  \argmin_{(\bu, \bv, \bw) \in \mathcal{H}_1 \times \mathcal{H}_2 \times \mathcal{H}_3} & \quad f(\bu) + g(\bv) + h(\bw)\\ \text{ subject to } & \quad  \mathfrak{L}_1\bu + \mathfrak{L}_2\bv + \mathfrak{L}_3\bw = \bb.
\end{aligned} \label{eqn:three_block}
\end{equation} Like the standard ADMM, the Davis-Yin scheme alternately updates the augmented Lagrangian in a Gauss-Seidel fashion: 
\begin{align*}
    \mathscr{L}_{\rho}(\bu, \bv, \bw, \bz) =& f(\bu) + g(\bv) + h(\bw) \\ &+ \frac{\rho}{2}\|\mathfrak{L}_1\bu + \mathfrak{L}_2\bv + \mathfrak{L}_3\bw - \bb - \rho^{-1}\bz\|_{\mathcal{H}_*}^2
\end{align*}
The Davis-Yin iterates bear similarities to the standard ADMM and the Alternating Minimization Algorithm \citep{Tseng:1991}, both of which are special cases: 
\begin{subequations}
\renewcommand{\theequation}{\theparentequation.\arabic{equation}}
\begin{align*}
  \bu^{(k+1)} &= \argmin_{\bu \in \mathcal{H}_1} f(\bu) + \langle \bz^{(k)}, \mathfrak{L}_1\bu\rangle\\
  \bv^{(k+1)} &= \argmin_{\bv \in \mathcal{H}_2} \mathscr{L}_{\rho}(\bu^{(k+1)}, \bv, \bw^{(k)}, \bz^{(k)}) \\
  \bw^{(k+1)} &= \argmin_{\bw \in \mathcal{H}_3} \mathscr{L}_{\rho}(\bu^{(k+1)}, \bv^{(k+1)}, \bw, \bz^{(k)}) \\
  \bz^{(k+1)} &= \bz^{(k)} + \rho(\mathfrak{L}_1\bu^{(k+1)} + \mathfrak{L}_2\bv^{(k+1)} + \mathfrak{L}_3\bw^{(k+1)} - \bb)
\end{align*}
\end{subequations}
Like the AMA, the Davis-Yin three block ADMM requires strong convexity of $f(\cdot)$. Unlike the standard ADMM, which allows $\rho$ to be arbitrarily chosen or even to vary over the course of the algorithm, both the AMA and Davis-Yin ADMM place additional constraints on $\rho$ to ensure convergence.

\section{Operator-Splitting Methods for Convex Bi-Clustering} \label{sec:algs}
In this section, we present three operator-splitting methods for convex co-clustering. For simplicity of exposition, we focus only on the bi-clustering (second-order tensor) case in this section and suppress the fusion weights $\bw, \tilde{\bw}$. The bi-clustering problem \eqref{eqn:bclust} then becomes 
\begin{equation}
\begin{aligned}
\hat{\bU} &= \argmin_{\bU \in \R^{n \times p}} \frac{1}{2}\|\bX - \bU\|_F^2 +\lambda \left(\|\bD_{\text{row}}\bU\|_{\text{row}, q} + \|\bU\bD_{\text{col}}\|_{\text{col}, q}\right)
\end{aligned} \label{eqn:bclust2}
\end{equation} where $\bD_{\text{row}}$ and $\bD_{\text{col}}$ are directed difference matrices implied by the (non-zero) fusion weights and $\|\cdot\|_{\text{row}, q}$ and $\|\cdot\|_{\text{col}, q}$ are the sums of the row- and column-wise $\ell_q$-norms of a matrix. The extension of our methods to to tensors of arbitrary order is natural and discussed in more detail in Section \ref{sec:tensor}.

\subsection{Alternating Direction Method of Multipliers}
Because the convex bi-clustering problem \eqref{eqn:bclust2} has three terms, it is not immediately obvious how a standard two-block ADMM can be applied to this problem. However, if we take $\mathcal{H}_1 = \R^{n \times p}$ to be the natural space of the primal variable and $\mathcal{H}_2 = \mathcal{H}_{\text{row}} \times \mathcal{H}_{\text{col}}$ to be the Cartesian product of the row- and column-edge difference spaces, we can obtain the ADMM updates:
\begin{align*}
  \bU^{(k+1)} &= \argmin_{\bU \in \R^{n \times p}} \frac{1}{2}\|\bX - \bU\|_2^2 + \frac{\rho}{2}\left\|\bD_{\text{row}}\bU - \bV^{(k)}_{\text{row}} + \rho^{-1}\bZ^{(k)}_{\text{row}}\right\|_F^2 \\ &\qquad +  \frac{\rho}{2}\left\|\bU\bD_{\text{col}} - \bV^{(k)}_{\text{col}} + \rho^{-1}\bZ^{(k)}_{\text{col}}\right\|_F^2 \\
  \begin{pmatrix} \bV^{(k+1)}_{\text{row}} \\ \bV^{(k+1)}_{\text{col}} \end{pmatrix} &= \begin{pmatrix} \prox_{\lambda / \rho \, \|\cdot\|_{\text{row}, q}}(\bD_{\text{row}}\bU^{(k+1)} + \rho^{-1}\bZ^{(k)}_{\text{row}}) \\
  \prox_{\lambda / \rho \, \|\cdot\|_{\text{col}, q}}(\bU^{(k+1)}\bD_{\text{col}} + \rho^{-1}\bZ^{(k)}_{\text{col}}) \end{pmatrix}  \\
  \begin{pmatrix} \bZ^{(k+1)}_{\text{row}} \\ \bZ^{(k+1)}_{\text{col}} \end{pmatrix} &= \begin{pmatrix} \bZ^{(k)}_{\text{row}} + \rho(\bD_{\text{row}}\bU^{(k+1)} - \bV^{(k+1)}_{\text{row}}) \\ \bZ^{(k)}_{\text{col}} + \rho(\bU^{(k+1)}\bD_{\text{col}} - \bV^{(k+1)}_{\text{col}}) \end{pmatrix}  
\end{align*}
The $\bV$- and $\bZ$-updates are straightforward to derive and parallel those of the ADMM for the convex clustering problem \citep[Appendix A.1]{Chi:2015,Weylandt:2019}. Additionally, we note that, due to the separable Cartesian structure of $\mathcal{H}_2$, the $\bV$ and $\bZ$-updates can be applied separately, and potentially in parallel, for the row and column blocks. As we will see, this separable structure holds for all algorithms considered in this paper and for tensors of arbitrary rank.

The $\bU$-update is somewhat more difficult as it requires us to solve 
\begin{equation*}
\begin{aligned}
&\bX + \rho\bD_{\text{row}}^T(\bV^{(k)}_{\text{row}} - \rho^{-1}\bZ^{(k)}_{\text{row}}) + \rho(\bV^{(k)}_{\text{col}} - \rho^{-1}\bZ^{(k)}_{\text{col}})\bD_{\text{col}}^T \\ &\quad = \bU + \rho\bD_{\text{row}}^T\bD_{\text{row}}\bU + \rho \bU\bD_{\text{col}}\bD_{\text{col}}^T.
\end{aligned} \label{eqn:sylvester}
\end{equation*}
This is a Sylvester equation with coefficient matrices $\frac{1}{2}\bI + \rho \bD_{\text{row}}^T\bD_{\text{row}}$ and $\frac{1}{2}\bI + \rho \bD_{\text{col}}\bD_{\text{col}}^T$ and can be solved using standard numerical algorithms \citep{Bartels:1972,Golub:1979}. Typically, the most expensive step in solving a Sylvester equation is either a Schur or Hessenberg decomposition of the coefficient matrices, both of which scale cubically with the size of the coefficient matrices. Since the coefficient matrices are fixed by the problem structure and do not vary between iterations, this factorization can be cached and amortized over the ADMM iterates, though, as we will see below, it is often more efficient to modify the $\bU$-update to avoid the Sylvester equation entirely.

\subsection{Generalized ADMM}

To avoid the Sylvester equation in the $\bU$-update, we turn to the Generalized ADMM framework of \citet{Deng:2016} and augment the $\bU$-update with the quadratic operator $\mathfrak{A}(\bU) = \frac{1}{2}(\alpha\|\bU\|_F^2 - \rho\|\mathfrak{L}_1\bU\|_{\mathcal{H}_*}^2)$. The constant $\alpha$ must be chosen to ensure $\mathfrak{A}$ is positive-definite so that sub-problem \eqref{eqn:gadmm_1} remains convex. The smallest valid $\alpha$ depends on $\rho$ and the operator norm of $\mathfrak{L}_1$ which can be bounded above by 
\[\|\mathfrak{L}_1\|_{\mathcal{H}_1 \to \mathcal{H}_*} \leq \sigma_{\max}(\bD_{\text{row}}) + \sigma_{\max}(\bD_{\text{col}})\]
where $\sigma_{\max}(\cdot)$ is the maximum singular value of a matrix. To avoid expensive singular value calculations, an upper bound based on the maximum degree of any vertex in the graph can be used instead \citep[Section 4.1]{Anderson:1985,Chi:2015}.

With this augmentation, the $\bU$-update \eqref{eqn:gadmm_1} becomes:
\begin{align*}
  \bU^{(k+1)} &= \left(\alpha \bU^{(k)} + \bX + \rho\bD_{\text{row}}^T(\bV^{(k)} - \rho^{-1}\bZ^{(k)}_{\text{row}} - \bD_{\text{row}}\bU^{(k)}) \right. \\ &\left.\qquad+ \rho(\bV^{(k)}_{\text{col}} - \rho^{-1}\bZ^{(k)}_{\text{col}} - \bU^{(k)}\bD_{\text{col}})\bD_{\text{col}}^T \right)/(1+\alpha)
\end{align*}
We note that this can be interpreted as a weighted average of a standard update and the previous estimate of $\bU$. This update does not require solving a Sylvester equation and can be performed in quadratic, rather than cubic, time. As we will see in Section \ref{sec:experiments}, this update significantly decreases the computational cost of an iteration without significantly slowing the \emph{per iteration} convergence rate, resulting in much improved performance.

\subsection{Davis-Yin Splitting}

The three-block structure of the convex bi-clustering problem \eqref{eqn:bclust2} naturally lends itself to the Davis-Yin three-block ADMM, which has provable convergence due to the strongly convex squared Frobenius norm loss. We take $\mathcal{H}_1$ as before, but now split the row- and column-edge differences into two spaces ($\mathcal{H}_2 = \mathcal{H}_{\text{row}}$ and $\mathcal{H}_3 = \mathcal{H}_{\text{col}}$). The joint range space, which also contains the dual variable, is still the Cartesian product $\mathcal{H}_* = \mathcal{H}_2 \times \mathcal{H}_3$. The Davis-Yin iterates have the same $\bV$- and $\bZ$-updates as before, but yield a simpler $\bU$-update: 
\begin{align*}
  \bU^{(k+1)} &= \bX - \bD_{\text{row}}^T\bZ^{(k)}_{\text{row}} - \bZ^{(k)}_{\text{col}}\bD^T_{\text{col}}
\end{align*}
Unlike the ADMM, the Davis-Yin updates require $\rho$ to be less than $1 / \|\mathcal{L}_1\|_{\mathcal{H}_1 \to \mathcal{H}_*}$. The same bound used for $\alpha$ in the Generalized ADMM can be used here.

We note that, due to the ``Cartesian'' structure of the linear operators $\mathcal{L}_2\bV = (-\bV_{\text{row}}, \bzero_{\mathcal{H}_{\text{col}}})$ and $\mathcal{L}_3\bW = (\bzero_{\mathcal{H}_{\text{row}}}, -\bV_{\text{col}})$, the Davis-Yin updates have the same separable structure as the ADMM and Generalized ADMM updates. In fact, it is not hard to see that the Davis-Yin updates can also be derived as updates of the Alternating Minimization Algorithm (AMA) \citep{Tseng:1991}. The AMA is known to be equivalent to proximal gradient on the dual problem, which \citet{Chi:2018b} recommended to solve Problem \eqref{eqn:cclust}.

As \citet{Chi:2015} note for the convex clustering problem, Moreau's decomposition \citep{Moreau:1962} can be used to eliminate the $\bV$-update by replacing the proximal operator with a projection onto the dual norm ball, giving the simplified $\bZ$-updates 
\begin{align*}
  \bZ^{(k+1)}_{\text{row}} &= \proj_{\lambda / \rho \mathbb{B}_{\|\cdot\|^*_{\text{row}, q}}}(\bZ^{(k)} + \rho \bD_{\text{row}}\bU^{(k+1)}) \\
  \bZ^{(k+1)}_{\text{col}} &= \proj_{\lambda / \rho \mathbb{B}_{\|\cdot\|^*_{\text{col}, q}}}(\bZ^{(k)} + \rho \bD_{\text{col}}\bU^{(k+1)}).
\end{align*} 
This simplification lowers the computational cost of the Davis-Yin updates, but it does not improve their per iteration convergence which, as we will see in the next section, is not competive with the ADMM variants.

\section{Simulation Studies} \label{sec:experiments}
In this section, we compare the performance of the three algorithms considered on the presidential speeches data set ($n = 44$, $p = 75$) of \citet{Weylandt:2019} and log-transformed Level III RPKM gene expression levels from 438 breast cancer samples collected by The Cancer Genome Atlas (TCGA) ($n = 438$, $p = 353$)  \citep{TCGA:2012}. In addition to the three algorithms discussed above, we also compare to the \cobra algorithm of \citet{Chi:2017}, an application of the \emph{Dykstra-Like Proximal Algorithm} of \citet{Bauschke:2008} to Problem \eqref{eqn:bclust}, which works by solving alternating convex clustering problems \eqref{eqn:clust} on the rows and columns of $\bU$ until convergence. For each method, we compare a standard version and a version using the acceleration techniques proposed in \citet{Goldstein:2014}. (For \cobra, acceleration was applied to both the row and column sub-problems.)

We show results for the rotationally-invariant $q = 2$ penalty with $\lambda = 1\times 10^4$ for the presidents data and $\lambda = 1\times 10^6$ for the breast cancer data, though our results are similar for other penalty functions and values of $\lambda$. For the ADMM variants, we fixed $\rho = 1$. The Davis-Yin step size and the Generalized ADMM coefficient ($\alpha$) were both set to twice the maximum degree of the row- or column-indicidence graph. The default sparse Gaussian kernel weights of the \clustRviz package \citep{Weylandt:2019-pkg} are used for each data set.

\begin{figure*}[t]
  \centering
  \includegraphics[width=\textwidth]{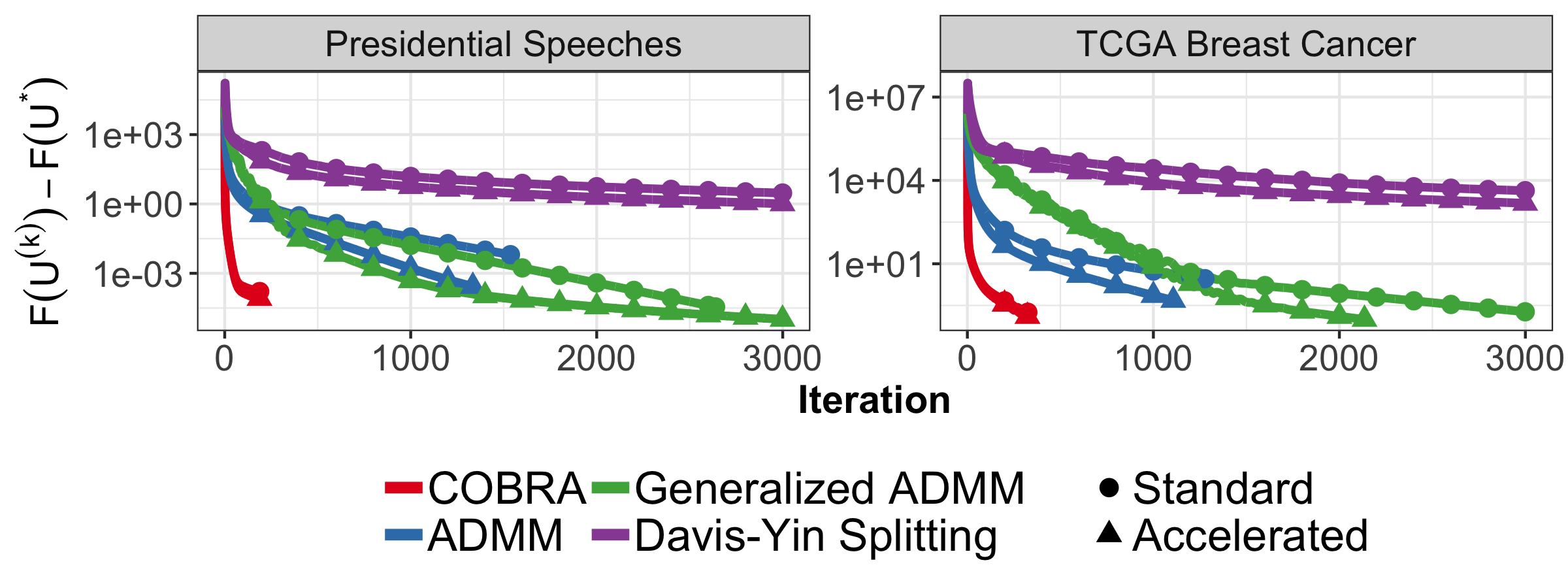}\vspace{-0.35in}
  \caption{Per iteration convergence of the ADMM, Generalized ADMM, Davis-Yin Splitting, and \cobra on the presidential speech and TCGA breast cancer data sets. \cobra clearly exhibits the fastest \emph{per iteration} convergence, appearing to exhibit super-linear convergence. The ADMM and Generalized ADMM have similar convergence rates, while the Davis-Yin iterates converge much more slowly.}
  \label{fig:iter_conv}
\end{figure*}

As can be seen in Figure \ref{fig:iter_conv}, the \cobra algorithm clearly has the fastest \emph{per iteration} convergence, rapidly converging to the optimal solution. (The slower convergence near optimality appears to be an artifact of the inexact solution of the sub-problems rather than an inherent behavior of the DLPA.) The ADMM and Generalized ADMM exhibit relatively fast convergence, while the Davis-Yin iterates are the slowest to converge. The standard ADMM appears to have slightly faster convergence than the Generalized ADMM near initialization, but the two methods appear to eventually attain the same convergence rate. Acceleration provides a minor, but consistent, improvement for all methods except \cobra, where the acceleration is applied within each sub-problem.

\begin{figure*}[t]
  \centering
  \includegraphics[width=\textwidth]{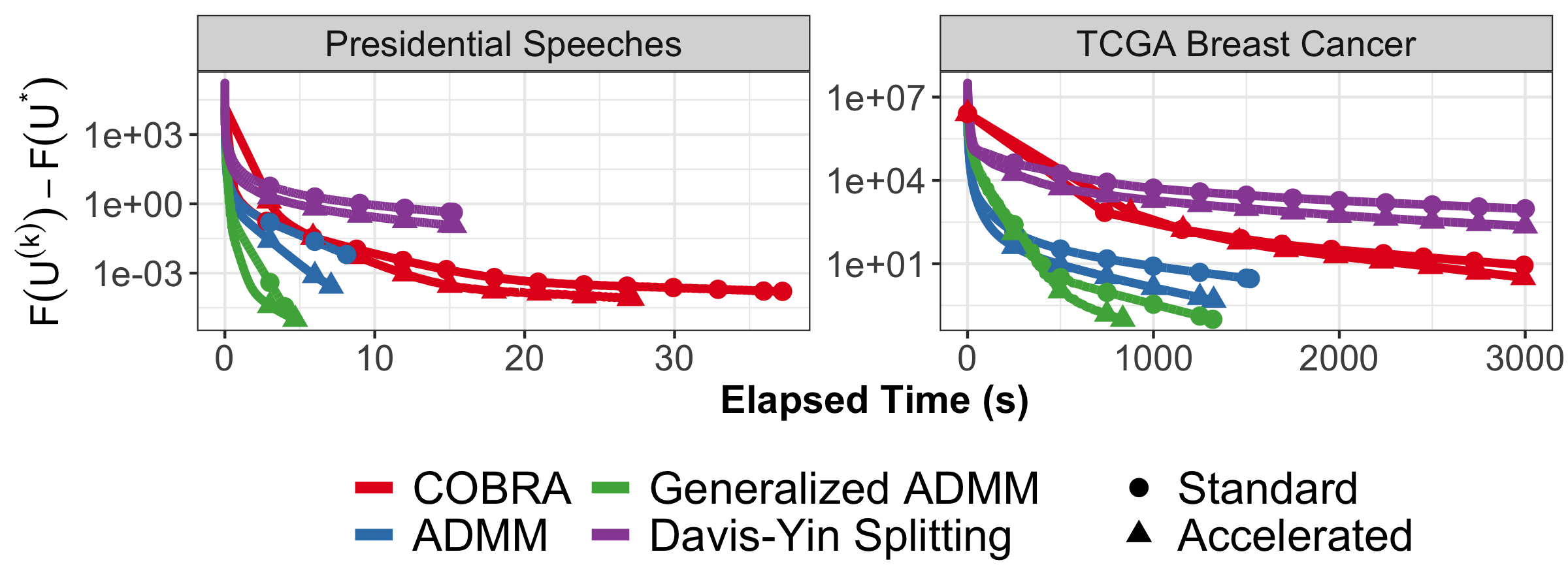}\vspace{-0.35in}
  \caption{Wall clock speed of the ADMM, Generalized ADMM, Davis-Yin Splitting, and \cobra on the presidential speech and TCGA data sets. Despite its rapid per iteration convergence, \cobra is less efficient than the ADMM and Generalized ADMM due to the complexity of its iterations.  Davis-Yin Splitting remains the least efficient algorithm considered for this problem.}
  \label{fig:iter_time}
\end{figure*}

The apparent advantage of \cobra disappears, however, when we instead consider total elapsed time, as shown in Figure \ref{fig:iter_time}, as the \cobra updates require solving a pair of expensive convex clustering sub-problems at each step. On a wall-clock basis, the Generalized ADMM performs the best for both problems, followed by the standard ADMM, \cobra, and Davis-Yin splitting in that order. 

Practically, the Generalized ADMM and Davis-Yin updates have essentially the same cost and are both three to four times faster per iteration than the standard ADMM with caching. Precise numerical results are given in Table \ref{tab:results}. These results suggest that our Generalized ADMM scheme attains the best of both worlds, achieving both the high per iteration convergence rate of a standard ADMM with the low per iteration computational cost of the Davis-Yin / AMA updates, consistent with our theoretical analysis in the next section.

\section{Complexity Analysis}
Both the ADMM and AMA are known to exhibit $O(1/K)$ convergence in general, \citep{He:2012,He:2015,Goldstein:2014}, but the ADMM obtains a superior convergence rate of $o(1/K)$ for strongly convex problems \citep{Davis:2017b}. This rate can be further improved to linear convergence if certain full-rank conditions are fulfilled by the problem-specific $\bD_{\text{row}}$ and $\bD_{\text{col}}$ matrices \citep{Deng:2016}. As \citet{Deng:2016} show, the Generalized ADMM achieves essentially the same convergence rate as the standard ADMM. Consequently, theory suggests that the two-block ADMM variants have superior per iteration convergence, consistent with our experimental results.

The computational cost of the Generalized ADMM and Davis-Yin  $\bU$-updates are essentially the same, both being dominated by matrix multiplications giving $\mathcal{O}(np\max(\#\text{rows}, \#\text{columns}))$ complexity, where $\#\text{rows}$ and $\#\text{columns}$ are the number of non-zero row and column fusion weights respectively. For sparse weighting schemes, we typically have $\#\text{rows} = \mathcal{O}(n)$ and $\#\text{columns} = \mathcal{O}(p)$, yielding an overall per iteration complexity of $\mathcal{O}(\max(np^2, n^2p))$ for the $\bU$-update. A naive implementation of the ADMM update requires $\mathcal{O}((\#\text{rows})^3 + (\#\text{columns})^3)$ to solve the Sylvester equation, but if the initial factorization is cached, the per iteration complexity is again reduced to $\mathcal{O}(\max(np^2, n^2p))$ under a sparse weight scheme.

To the best of our knowledge, a convergence rate for the DLPA, on which \cobra is based, has not been established. 
Experimentally, the DLPA exhibits very rapid convergence, consistent with known rates for Dykstra's alternating projections algorithm. Despite this, the high per iteration cost makes it impractical on larger problems.

\begin{table*}
\begin{center}
\begin{tabular}{lS[table-format=3.2]rrS[table-format=3.2]rr}
\toprule
\multirow{2}{*}{Method}& \multicolumn{3}{c}{\emph{Presidential Speeches}} & \multicolumn{3}{c}{\emph{TCGA Breast Cancer}} \\
& {Iter/sec} & {Total Time (s)} & {Total Iterations} & {Iter/sec} & {Total Time (s)} & {Total Iterations} \\
\midrule
Generalized ADMM & 663 & 3.98 & 2639 & 2.58 & 1318 & 3411\\ 
\phantom{AB} Accelerated & 645 & 4.69 & 3025 & 2.55 & 835 & 2135\\
Davis-Yin / AMA & 657 & \textsuperscript{\textdagger}15.23 & \textsuperscript{\textdagger}10000 & 2.60 & \textsuperscript{\textdagger}3850 &  \textsuperscript{\textdagger}10000\\
\phantom{AB} Accelerated & 653 & \textsuperscript{\textdagger}15.31 & \textsuperscript{\textdagger}10000 & 2.58 & \textsuperscript{\textdagger}3879 & \textsuperscript{\textdagger}10000\\
ADMM & 189 & 8.12 & 1536 & 0.84 & 1521 & 1277\\ 
\phantom{AB} Accelerated & 188 & 7.08 & 1334 & 0.84 & 1324 & 1107\\
\midrule
\cobra & 5.06 & 37.20 & 188 (29581 total) & 0.054 & 5994 & 325 (16519 total)\\
\phantom{AB} Accelerated & 6.89 & 27.15 & 187 (21362 total) & 0.065 & 4953 & 323 (13549 total)\\
\bottomrule
\end{tabular}
\end{center}\vspace{-0.15in}
\caption{Performance of the ADMM, Generalized ADMM, Davis-Yin Splitting, and \cobra on the presidential speech and TCGA data sets. For \cobra, the numbers in parentheses are the total number of sub-problem iterations taken. The $\dagger$ indicates that the method was stopped after failing to converge in 10,000 iterations.}
\label{tab:results}
\end{table*}

\section{Extensions to General Co-Clustering} \label{sec:tensor}

In Section \ref{sec:algs}, we restricted our attention to co-clustering tensors of order two, \emph{i.e.}, bi-clustering. In this section, we show how the ADMM, Generalized ADMM, and Davis-Yin (AMA) algorithms can be extended to the general problem of co-clustering order-$K$ tensors \eqref{eqn:cclust}. As before, we can re-express the co-clustering problem using directed difference tensors $\mathcal{D}_j$ as
\[\hat{\mathcal{U}} = \argmin_{\mathcal{U}} \frac{1}{2}\|\mathcal{X} - \mathcal{U}\|_F^2 + \lambda \sum_{j=1}^J\|\mathcal{U} \times_j \mathcal{D}_j\|_{j, q}\] where $\|\mathcal{X}\|_{j, q}$ is the sum of the $\ell_q$-norms of the vectorization of each slice of $\mathcal{X}$ along the $j$\textsuperscript{th} mode. For all three methods, the $\mathcal{V}$ and $\mathcal{Z}$-updates are straightforward extensions of the bi-clustering case: 
\begin{align*}
  \mathcal{V}^{(k+1)}_j &= \prox_{\lambda/\rho \|\cdot\|_{j, q}}\left(\mathcal{U}^{(k+1)} \times_j \mathcal{D}_j + \rho^{-1} \mathcal{Z}{(k)}_j\right) \\
  \mathcal{Z}^{(k+1)}_j &= \mathcal{Z}^{(k)}_j + \rho(\mathcal{U}^{(k+1)}\times_j \mathcal{D}_j - \mathcal{V}^{(k+1)}_j).
\end{align*}
As with the bi-clustering case, we have separate $\mathcal{V}$ and $\mathcal{Z}$ variables for each term in the fusion penalty (\emph{i.e.}, each mode of the tensor), each of which can be updated in parallel. For the standard ADMM, the $\mathcal{U}$-update requires solving the following tensor Sylvester equation, where $(\cdot)^{\mathcal{T}}$ denotes a transpose, with the direction implied by context:
\[\mathcal{X} + \rho\sum_{j=1}^J (\mathcal{V}^{(k)}_j - \rho^{-1}\mathcal{Z}^{(k)}_j) \times_j \mathcal{D}_j^{\mathcal{T}} = \mathcal{U} + \rho\sum_{j=1}^J \mathcal{U} \times_j \mathcal{D}_j \times_j \mathcal{D}_j^\mathcal{T}.\] As above, the $\mathcal{U}$-update can be solved explicitly for our other methods:
\begin{align*}
  \mathcal{U}^{(k+1)} &=\frac{\alpha}{1+\alpha}\mathcal{U}^{(k)} + \frac{\mathcal{X}}{1+\alpha} \tag{Generalized ADMM} \\ &\quad + \frac{\rho}{1+\alpha}\sum_{j=1}^J(\mathcal{V}^{(k)}_j - \rho^{-1}\mathcal{Z}^{(k)}_j - \mathcal{U}^{(k)}\times_j\mathcal{D}_j)\times_j\mathcal{D}_j^{\mathcal{T}}  \\
  \mathcal{U}^{(k+1)} &= \mathcal{X} -\sum_{j=1}^J \mathcal{Z}^{(k)}_j \times_j (\mathcal{D}_j)^{\mathcal{T}}. \tag{Davis-Yin / AMA}
\end{align*}
\section{Discussion}
We have introduced three operator-splitting methods for solving the convex bi-clustering and co-clustering problems: a standard ADMM, a Generalized ADMM, and a three-block ADMM based on Davis-Yin splitting. The Davis-Yin three-block ADMM was found to be equivalent to the AMA which is in turn equivalent to the dual projected gradient recommended by \citet{Chi:2018b}. The standard ADMM achieves the fastest per iteration convergence, but requires solving a Sylvester equation at each step. The Generalized ADMM avoids the expensive Sylvester equation while still maintaining the rapid convergence of the ADMM. Unlike \citet{Chi:2015}, we do not find that the AMA performs well, perhaps due to the more complex structure of the bi- and co-clustering problems. On the TCGA breast cancer data, the Generalized ADMM was able to find a better solution in three and a half minutes than the AMA could in over an hour. All three of our methods can be extended to co-cluster tensors of arbitrary order, with the advantages of the Generalized ADMM becoming even more pronounced as it avoids a tensor Sylvester equation.

Our ADMM and Generalized ADMM methods provide a unified and efficient computational approach to convex co-clustering of large tensors of arbitrary order, but several interesting methodological questions remain unanswered. In particular, the conditions under which the co-clustering solution path is purely agglomerative, enabling the construction of dendrogram-based visualizations of the kind proposed for convex clustering by \citet{Weylandt:2019}, remain unknown. While not discussed here, our algorithms can naturally be extended for missing data handling and cross-validation, using a straight-forward extension of the Majorization-Minimization (MM) imputation scheme proposed by \citet{Chi:2017}. Our algorithms make convex bi- and co-clustering practical for large structured data sets and we anticipate they will encourage adoption and additional study of these useful techniques.

\section{References}
\printbibliography[heading=none]
\clearpage
\onecolumn

\appendix

\section{Step Size Selection}
The Generalized ADMM requires that the quadratic operator $\mathfrak{A}(\bU) = \alpha\|\bU\|_F^2 - \rho\|\mathfrak{L}_1\bU\|_{\mathcal{H}^*}^2$ be positive-definite, which holds if $\alpha / \rho$ is greater than the operator norm of $\mathfrak{L}_1$. Similarly, the Davis-Yin ADMM requires that the step size $\rho^{-1}$ be greater the operator norm of $\mathfrak{L}_1$. (See the related discussion in Section 4.2 of \citet{Chi:2015}.) While this operator norm is somewhat inconvenient to compute, an upper bound is easily found: 
\begin{align*}
\|\mathfrak{L}_1\|_{\mathcal{H}_1 \to \mathcal{H}_*} &= \sup_{\substack{\bU \in \mathcal{H}_1\\ \|\bU\|_{\mathcal{H}_1} \leq 1}} \|\mathfrak{L}_1\bU\|_{\mathcal{H}_*} \\
&= \sup_{\substack{\bU \in \mathcal{H}_1\\ \|\bU\|_{\mathcal{H}_1} \leq 1}} \sqrt{\|\bD_{\text{row}}\bU\|_F^2 + \|\bU\bD_{\text{col}}\|_F^2} \\
&\leq \sup_{\substack{\bU \in \mathcal{H}_1\\ \|\bU\|_{\mathcal{H}_1} \leq 1}} \|\bD_{\text{row}}\bU\|_F + \|\bU\bD_{\text{col}}\|_F \\ 
&\leq \sup_{\substack{\bU \in \mathcal{H}_1\\ \|\bU\|_{\mathcal{H}_1} \leq 1}} \|\bD_{\text{row}}\bU\|_F + \sup_{\substack{\bU \in \mathcal{H}_1\\ \|\bU\|_{\mathcal{H}_1} \leq 1}} \|\bU\bD_{\text{col}}\|_F \\ 
&= \sigma_{\max}(\bD_{\text{row}}) + \sigma_{\max}(\bD_{\text{col}})
\end{align*}
where $\sigma_{\max}(\cdot)$ is the maximum singular value of a matrix. To avoid a potentially expensive SVD for large problems, we can instead take advantage of the special structure of $\bD_{\text{row}}$ and $\bD_{\text{col}}$ using known results about the eigenstructure of a graph Laplacian \citep{Anderson:1985} which imply \[\sigma_{\max}(\bD_{\text{row}}) \leq \max_{(i, j) \in \mathcal{E}_{\text{row}}} \text{degree}(i) + \text{degree}(j) \leq 2 \max_{i \in \mathcal{E}_{\text{row}}} \text{degree}(i) \quad \text{ and } \quad \sigma_{\max}(\bD_{\text{col}}) \leq \max_{(i, j) \in \mathcal{E}_{\text{col}}} \text{degree}(i) + \text{degree}(j) \leq 2 \max_{i \in \mathcal{E}_{\text{col}}} \text{degree}(i) \] where the first maximum is taken over the pairs of connected vertices and the second is taken over all vertices. This bound can be computed with minimal effort and, in our experience, is sufficiently tight for most problems.

\section{Detailed Derivations}
In this section, we provide detailed derivations of the ADMM, Generalized ADMM, and Davis-Yin algorithms for convex clustering discussed above. Unless otherwise stated, our notation is the same as that used in the main body of the paper. Given the large number of Hilbert spaces, norms, and inner products used in these derivations, we err on the side of explicitness rather than concision in our notation. 
\subsection{ADMM}
The ADMM updates are
\begin{align*}
  \bU^{(k+1)} &= \argmin_{\bU \in \mathcal{H}_1} f(\bU) + \langle \bZ^{(k)}, \mathfrak{L}_1\bU + \mathfrak{L}_2\bV^{(k)} - \bb\rangle + \frac{\rho}{2}\left\|\mathfrak{L}_1\bU + \mathfrak{L}_2\bV^{(k)} - \bb\right\|_{\mathcal{H}_*}^2 \\
  \bV^{(k+1)} &= \argmin_{\bV \in \mathcal{H}_2} g(\bV) + \langle \bZ^{(k)}, \mathfrak{L}_1\bU^{(k+1)} + \mathfrak{L}_2\bV - \bb\rangle + \frac{\rho}{2}\left\|\mathfrak{L}_1\bU^{(k+1)} + \mathfrak{L}_2\bV - \bb\right\|_{\mathcal{H}_*}^2 \\
  \bZ^{(k+1)} &= \bZ^{(k)} + \rho(\mathfrak{L}_1\bU^{(k+1)} + \mathfrak{L}_2\bV^{(k+1)} - \bb).
\end{align*}
Rescaling the dual variable by $\rho^{-1}$ ($\bZ \to \rho^{-1}\bZ$), these can be simplified to
\begin{align*}
  \bU^{(k+1)} &= \argmin_{\bU \in \mathcal{H}_1} f(\bU) + \frac{\rho}{2}\left\|\mathfrak{L}_1\bU + \mathfrak{L}_2\bV^{(k)} - \bb + \bZ^{(k)}\right\|_{\mathcal{H}_*}^2 \\
  \bV^{(k+1)} &= \argmin_{\bV \in \mathcal{H}_2} g(\bV) +  \frac{\rho}{2}\left\|\mathfrak{L}_1\bU^{(k+1)} + \mathfrak{L}_2\bV - \bb + \bZ^{(k)}\right\|_{\mathcal{H}_*}^2 \\
  \bZ^{(k+1)} &= \bZ^{(k)} + \mathfrak{L}_1\bU^{(k+1)} + \mathfrak{L}_2\bV^{(k+1)} - \bb.
\end{align*}
For the convex bi-clustering problem \eqref{eqn:bclust2}, we take: 
\begin{itemize}
  \item $\mathcal{H}_1 = \R^{n \times p}$ equipped with the Frobenius norm and inner product
  \item $\mathcal{H}_{\text{row}} = \R^{|\mathcal{E}_{\text{row}}| \times p}$ and $\mathcal{H}_{\text{col}} = \R^{n \times |\mathcal{E}_{\text{col}}|}$ both equipped with the Frobenius norm and inner product
  \item $\mathcal{H}_2 = \mathcal{H}_* = \mathcal{H}_{\text{row}} \times \mathcal{H}_{\text{col}}$ equipped with the inner product \[\langle (\bA_{\text{row}}, \bA_2), (\bB_{\text{row}}, \bB_{\text{col}})\rangle_{\mathcal{H}_{*}} = \langle \bA_{\text{row}}, \bB_{\text{row}}\rangle_{\mathcal{H}_{\text{row}}} + \langle \bA_{\text{col}}, \bB_{\text{col}}\rangle_{\mathcal{H}_{\text{col}}}\] and the norm \[\|(\bA_{\text{row}}, \bA_{\text{col}})\|_{\mathcal{H}_{*}} = \sqrt{\|\bA_{\text{row}}\|_{\mathcal{H}_{\text{row}}}^2 + \|\bA_{\text{col}}\|_{\mathcal{H}_{\text{col}}}^2}\]
  \item $\mathfrak{L}_1: \mathcal{H}_1 \to \mathcal{H}_{*}$ given by $\mathfrak{L}_1\bU = (\bD_{\text{row}}\bU, \bU\bD_{\text{col}})$
  \item $\mathfrak{L}_2: \mathcal{H}_2 \to \mathcal{H}_{*}$ given by $\mathfrak{L}_2(\bV_{\text{row}}, \bV_{\text{col}}) = -(\bV_{\text{row}}, \bV_{\text{col}}) $
  \item $\bb = \bzero_{\mathcal{H}_*} = (\bzero_{\mathcal{H}_{\text{row}}},\bzero_{\mathcal{H}_{\text{col}}}) \in \mathcal{H}_{*}$
  \item $f(\bU) = \frac{1}{2}\|\bX - \bU\|_F^2$
  \item $g((\bV_{\text{row}}, \bV_{\text{col}})) = \lambda \|\bV_{\text{row}}\|_{q, 1} + \lambda \|\bV_{\text{col}}\|_{1, q}$
\end{itemize}
First, we consider the $\bU$-update: 
\begin{align*}
\bU^{(k+1)} &= \argmin_{\bU \in \R^{n \times p}} \frac{1}{2}\|\bX - \bU\|_F^2 + \frac{\rho}{2}\left\|\begin{pmatrix} \bD_{\text{row}}\bU \\ \bU\bD_{\text{col}}\end{pmatrix} + \begin{pmatrix} -\bV_{\text{row}}^{(k)} \\ -\bV_{\text{col}}^{(k)}\end{pmatrix} + \begin{pmatrix} \bZ^{(k)}_{\text{row}} \\ \bZ^{(k)}_{\text{col}} \end{pmatrix}\right\|^2_{\mathcal{H}_*}  \\
&= \argmin_{\bU \in \R^{n \times p}} \frac{1}{2}\|\bU - \bX\|_F^2 + \frac{\rho}{2}\left\|\bD_{\text{row}}\bU - \bV^{(k)}_{\text{row}} + \bZ^{(k)}_{\text{row}}\right\|_{F}^2 + \frac{\rho}{2}\left\|\bU\bD_{\text{col}} - \bV^{(k)}_{\text{col}} + \bZ^{(k)}_{\text{col}}\right\|_{F}^2
\end{align*}
using the separability of the squared $\mathcal{H}_*$ norm ($\|\cdot\|_{\mathcal{H}_*}^2$). This is fully smooth and so we take the gradient with respect to $\bU$ to obtain the stationarity conditions: 
\[\bzero_{\mathcal{H}_1} = \bU - \bX + \rho\left(\bD_{\text{row}}^T(\bD_{\text{row}}\bU -\bV^{(k)}_{\text{row}} + \bZ^{(k)}_{\text{row}})\right) + \rho\left((\bU\bD_{\text{col}} - \bV^{(k)}_{\text{col}} + \bZ^{(k)}_{\text{col}})\bD_{\text{col}}^T\right)\]
using the identity\footnote{See Equation (119) in the Matrix Cookbook: \url{https://www.math.uwaterloo.ca/~hwolkowi/matrixcookbook.pdf}.}
\[\frac{\partial}{\partial \bX}\|\bA\bX\bB + \bC\|_F^2 = 2\bA^T[\bA\bX\bB + \bC]\bB^T.\] Solving for $\bU$, we obtain
\[\bX + \rho\bD^T_{\text{row}}(\bV^{(k)}_{\text{row}} - \bZ^{(k)}_{\text{row}}) + \rho(\bV^{(k)}_{\text{col}} + \bZ^{(k)}_{\text{col}})\bD^T_{\text{col}} = \bU + \rho\bD^T_{\text{row}}\bD_{\text{row}}\bU + \rho \bU\bD_{\text{col}}\bD_{\text{col}}^T\]

The $\bV$-updates are straight-forward to derive using the separable structure of $\mathcal{H}_2 = \mathcal{H}_{\text{row}} \times \mathcal{H}_{\text{col}}$:
\begin{align*}
  \begin{pmatrix} \bV^{(k+1)}_{\text{row}} \\ \bV^{(k+1)}_{\text{col}} \end{pmatrix} &= \argmin_{(\bV_{\text{row}}, \bV_{\text{col}}) \in \mathcal{H}_{\text{row}} \times \mathcal{H}_{\text{col}}} \lambda \|\bV_{\text{row}}\|_{q, 1} + \lambda \|\bV_{\text{row}}\|_{q, 1} + \frac{\rho}{2}\left\|\begin{pmatrix} \bD_{\text{row}}\bU \\ \bU\bD_{\text{col}} \end{pmatrix} + \begin{pmatrix} -\bV^{(k)}_{\text{row}} \\ -\bV^{(k)}_{\text{col}} \end{pmatrix} + \begin{pmatrix} \bzero_{\mathcal{H}_{\text{row}}} \\ \bzero_{\mathcal{H}_{\text{col}}} \end{pmatrix} + \begin{pmatrix} \bZ^{(k)}_{\text{row}} \\ \bZ^{(k)}_{\text{col}} \end{pmatrix} \right\|_{\mathcal{H}_{\text{row}} \times \mathcal{H}_{\text{col}}}^2  \\
  &= \begin{pmatrix} \argmin_{\bV_{\text{row}} \in \mathcal{H}_{\text{row}}} \lambda \|\bV_{\text{row}}\|_{\text{row}, q} + \frac{\rho}{2}\|\bD_{\text{row}}\bU^{(k+1)} - \bV_{\text{row}} + \bZ^{(k)}_{\text{row}}\|_F^2 \\ \argmin_{\bV_{\text{col}} \in \mathcal{H}_{\text{col}}} \lambda \|\bV_{\text{col}}\|_{\text{col}, q} + \frac{\rho}{2}\|\bU^{(k+1)}\bD_{\text{col}} - \bV_{\text{col}} + \bZ^{(k)}_{\text{col}}\|_F^2 \end{pmatrix}  \\
  &= \begin{pmatrix} \argmin_{\bV_{\text{row}} \in \mathcal{H}_{\text{row}}} \frac{\lambda}{\rho} \|\bV_{\text{row}}\|_{\text{row}, q} + \frac{1}{2}\|\bV_{\text{row}} -(\bD_{\text{row}}\bU^{(k+1)} + \bZ^{(k)}_{\text{row}})\|_F^2 \\ \argmin_{\bV_{\text{col}} \in \mathcal{H}_{\text{col}}} \frac{\lambda}{2} \|\bV_{\text{col}}\|_{\text{col}, q} + \frac{1}{2}\|\bV_{\text{col}} - (\bU^{(k+1)}\bD_{\text{col}} + \bZ^{(k)}_{\text{col}})\|_F^2 \end{pmatrix}  \\
  &= \begin{pmatrix} \prox_{\lambda / \rho \, \|\cdot\|_{\text{row}, q}}(\bD_{\text{row}}\bU^{(k+1)} + \bZ^{(k)}_{\text{row}}) \\
  \prox_{\lambda / \rho \, \|\cdot\|_{\text{col}, q}}(\bU^{(k+1)}\bD_{\text{col}} + \bZ^{(k)}_{\text{col}}) \end{pmatrix}
\end{align*}
If fusion weights are included, then they are reflected in the row- and column-wise proximal operators. Finally, the $\bZ$-updates are standard: 
\begin{align*}
  \begin{pmatrix} \bZ^{(k+1)}_{\text{row}} \\ \bZ^{(k+1)}_{\text{col}} \end{pmatrix} &= \begin{pmatrix} \bZ^{(k)}_{\text{row}} + \bD_{\text{row}}\bU^{(k+1)} - \bV^{(k+1)} \\ \bZ^{(k)}_{\text{col}} + \bU^{(k+1)}\bD_{\text{col}} - \bV^{(k+1)} \end{pmatrix}  
\end{align*}
If we unscale $\bZ \to \rho^{-1}\bZ$, we recover the updates given in the main body of this note.

\subsection{Generalized ADMM}
To avoid the Sylvester equation, we recommend use of a Generalized ADMM, where the $\bU$-update is augmented with the quadratic operator $\mathfrak{A}(\bU - \bU^{(k)}) = \frac{\alpha}{2}\|(\bU - \bU^{(k)})\|_F^2 - \frac{\rho}{2}\|\mathfrak{L}_1(\bU - \bU^{(k)})\|_F^2$ where $\alpha$ is fixed sufficiently large to ensure positive-definiteness. This augmentation gives: 
\begin{align*}
\bU^{(k+1)} &= \argmin_{\bU \in \R^{n \times p}} \frac{1}{2}\|\bU - \bX\|_F^2 + \frac{\rho}{2}\|\bD_{\text{row}}\bU - \bV^{(k)}_{\text{row}} + \bZ^{(k)}_{\text{row}}\|_F^2 + \frac{\rho}{2}\|\bD_{\text{row}}\bU - \bV^{(k)}_{\text{row}} + \bZ^{(k)}_{\text{row}}\|_F^2 \\ &\qquad \quad + \alpha\|\bU - \bU^{(k)}\|_F^2 - \rho\|\bD_{\text{row}}\bU - \bD_{\text{row}}\bU^{(k)}\|_F^2 + \rho\|\bU\bD_{\text{col}} - \bU^{(k)}\bD_{\text{col}}\|_F^2
\end{align*}
As before, we take gradients with respect to $\bU$ to obtain the stationarity condition: 
\begin{align*}
  \bzero_{\mathcal{H}_1} &= \bU - \bX + \rho\bD_{\text{row}}^T\bD_{\text{row}}\bU - \rho\bD_{\text{row}}^T(\bV^{(k)}_{\text{row}} - \bZ^{(k)}_{\text{row}}) + \rho\bU\bD_{\text{col}}\bD_{\text{col}}^T - \rho(\bV^{(k)}_{\text{col}} - \bZ^{(k)}_{\text{col}})\bD_{\text{col}}^T \\&\quad + \alpha (\bU - \bU^{(k)}) - \rho\bD_{\text{row}}^T(\bD_{\text{row}}\bU - \bD_{\text{row}}\bU^{(k)}) - \rho(\bU\bD_{\text{col}} - \bU^{(k)}\bD_{\text{col}})\bD_{\text{col}}^T   
\end{align*}
Simplifying, we obtain:
\[(1+\alpha)\bU = \bX + \rho\bD_{\text{row}}^T(\bV^{(k)}_{\text{row}} - \bZ^{(k)}_{\text{row}}) + \rho(\bV^{(k)}_{\text{col}} - \bZ^{(k)}_{\text{col}})\bD_{\text{col}}^T + \alpha \bU^{(k)} - \rho\bD^T_{\text{row}}\bD_{\text{row}}\bU^{(k)} - \rho\bU^{(k)}\bD_{\text{col}}\bD_{\text{col}}^T \]
The $\bV$ and $\bZ$-updates are unchanged.
\subsection{Davis-Yin}
The Davis-Yin updates are given by: 
\begin{align*}
  \bU^{(k+1)} &= \argmin_{\bU \in \mathcal{H}_1} f(\bu) + \langle \bZ^{(k)}, \mathfrak{L}_1\bU\rangle\\
  \bV^{(k+1)} &= \argmin_{\bV \in \mathcal{H}_2} g(\bV) + \frac{\rho}{2}\left\|\mathfrak{L}_1\bU^{(k+1)} + \mathfrak{L}_2\bV + \mathfrak{L}_3\bW^{(k)} - \bb + \rho^{-1}\bZ^{(k)})\right\|_{\mathcal{H}^*}^2 \\
  \bW^{(k+1)} &= \argmin_{\bw \in \mathcal{H}_3} h(\bW) + \frac{\rho}{2}\left\|\mathfrak{L}_1\bU^{(k+1)} + \mathfrak{L}_2\bV^{(k+1)} + \mathfrak{L}_3\bW - \bb + \rho^{-1}\bZ^{(k)})\right\|_{\mathcal{H}^*}^2 \\
  \bZ^{(k+1)} &= \bZ^{(k)} + \rho(\mathfrak{L}_1\bU^{(k+1)} + \mathfrak{L}_2\bV^{(k+1)} + \mathfrak{L}_3\bW^{(k+1)} - \bb)
\end{align*}
For the convex bi-clustering problem \eqref{eqn:bclust2}, we take:
\begin{itemize}
  \item $\mathcal{H}_1 = \R^{n \times p}$ equipped with the Frobenius norm and inner product
  \item $\mathcal{H}_2 = \mathcal{H}_{\text{row}} = \R^{|\mathcal{E}_{\text{row}}| \times p}$ equipped with the Frobenius norm and inner product
  \item $\mathcal{H}_3 = \mathcal{H}_{\text{col}} = \R^{n \times |\mathcal{E}_{\text{col}}|}$ equipped with the Frobenius norm and inner product
  \item $\mathcal{H}_{*} = \mathcal{H}_2 \times \mathcal{H}_3$ equipped with the inner product \[\langle (\bA_1, \bA_2), (\bB_1, \bB_2)\rangle_{\mathcal{H}_{*}} = \langle \bA_1, \bB_1\rangle_{\mathcal{H}_2} + \langle \bA_2, \bB_2\rangle_{\mathcal{H}_3}\] and the norm \[\|(\bA_1, \bA_2)\|_{\mathcal{H}_{*}} = \sqrt{\|\bA_1\|_{\mathcal{H}_2}^2 + \|\bA_2\|_{\mathcal{H}_3}^2}\]
  \item $\mathfrak{L}_1: \mathcal{H}_1 \to \mathcal{H}_{*}$ given by $\mathfrak{L}_1\bU = (\bD_{\text{row}}\bU, \bU\bD_{\text{col}})$
  \item $\mathfrak{L}_2: \mathcal{H}_2 \to \mathcal{H}_{*}$ given by $\mathfrak{L}_2\bV = (-\bV, \bzero_{\mathcal{H}_3})$
  \item $\mathfrak{L}_3: \mathcal{H}_3 \to \mathcal{H}_{*}$ given by $\mathfrak{L}_3\bW = (\bzero_{\mathcal{H}_2}, -\bW)$
  \item $\bb = \bzero_{\mathcal{H}_*} = (\bzero_{\mathcal{H}_2},\bzero_{\mathcal{H}_3}) \in \mathcal{H}_{*}$
  \item $f_1(\bU) = \frac{1}{2}\|\bX - \bU\|_F^2$
  \item $f_2(\bV) = \lambda \|\bV\|_{\text{row}, q}$
  \item $f_3(\bW) = \lambda \|\bW\|_{\text{col}, q}$
\end{itemize}
First, we consider the $\bU$-update:
\begin{align*}
\argmin_{\bU \in \mathcal{H}_1} \frac{1}{2}\|\bX - \bU\|_F^2 + \langle \bZ^{(k)}, \mathfrak{L}_1\bU\rangle_{\mathcal{H}_{*}}
&= \argmin_{\bU \in \mathcal{H}_1} \frac{1}{2}\|\bX - \bU\|_F^2 + \langle \bZ^{(k)}, (\bD_{\text{row}}\bU, \bU\bD_{\text{col}})\rangle_{\mathcal{H}_{*}} \\
&= \argmin_{\bU \in \mathcal{H}_1} \frac{1}{2}\|\bX - \bU\|_F^2 +\langle \bZ^{(k)}_{\text{row}}, \bD_{\text{row}}\bU\rangle_{\mathcal{H}_2} + \langle \bZ^{(k)}_{\text{col}}, \bU\bD_{\text{col}}\rangle_{\mathcal{H}_3} \\
&= \argmin_{\bU \in \mathcal{H}_1} \frac{1}{2}\|\bX - \bU\|_F^2 + \Tr((\bZ^{(k)}_{\text{row}})^T\bD_{\text{row}}\bU) + \Tr((\bZ^{(k)}_{\text{col}})^T\bU\bD_{\text{col}})
\end{align*}
This is fully smooth, so we take the gradient with respect to $\bU$, using the identity\footnote{See Equation (101) of the Matrix Cookbook: \url{https://www.math.uwaterloo.ca/~hwolkowi/matrixcookbook.pdf}.} 
\[\frac{\partial}{\partial \bX}\Tr(\bA\bX\bB) = \bA^T\bB^T \implies \frac{\partial}{\partial \bX}\Tr(\bA\bX) = \bA^T, \]
to obtain the stationarity condition
\[\bzero_{\mathcal{H}_1} = -(\bX - \bU) + \bD^T_{\text{row}}\bZ^{(k)}_{\text{row}} - \bZ^{(k)}_{\text{col}}\bD_{\text{col}}^T.\]
Solving for $\bU$, we obtain the Davis-Yin $\bU$-update:
\[\bU^{(k+1)} = \bX - \bD^T_{\text{row}}\bZ^{(k)}_{\text{row}} - \bZ^{(k)}_{\text{col}}\bD_{\text{col}}^T.\]
The $\bV_{\text{row}}$-update (\emph{i.e.}, the $\bV$-update above) is given by
\begin{align*}
&\argmin_{\bV \in \mathcal{H}_{\text{row}}} \lambda\|\bV\|_{\text{row}, q} + \frac{\rho}{2}\|\mathfrak{L}_1\bU^{(k+1)} + \mathfrak{L}_2\bV + \mathfrak{L}_3\bW^{(k)} - \bzero_{\mathcal{H}_*} - \rho^{-1}\bZ^{(k)}\|^2_{\mathcal{H}_{*}} \\
&\argmin_{\bV \in \mathcal{H}_{\text{row}}} \frac{\lambda}{\rho}\|\bV\|_{\text{row}, q} + \frac{1}{2}\|(\bD_{\text{row}}\bU^{(k+1)}, \bU^{(k+1)}\bD_{\text{col}}) + (-\bV, \bzero_{\mathcal{H}_{\text{col}}}) + (\bzero_{\mathcal{H}_{\text{row}}}, -\bW^{(k)}) + (\rho^{-1}\bZ^{(k)}_{\text{row}}, \rho^{-1}\bZ^{(k)}_{\text{col}})\|^2_{\mathcal{H}_{*}} \\
&\argmin_{\bV \in \mathcal{H}_{\text{row}}} \frac{\lambda}{\rho}\|\bV\|_{\text{row}, q} + \frac{1}{2}\left\|(\bD_{\text{row}}\bU^{(k+1)} - \bV +\rho^{-1}\bZ^{(k)}_{\text{row}}, \bU^{(k+1)}\bD_{\text{col}} - \bW^{(k)} +\rho^{-1}\bZ^{(k)}_{\text{col}}) \right\|_{\mathcal{H}_*}^2
\end{align*}
Using the separability of the squared $\mathcal{H}_*$-norm, we see that the $\mathcal{H}_3$ component of the second term does not depend on $\bV$ and hence the above problem is equivalent to 
\[\bV^{(k+1)}_{\text{row}} = \argmin_{\bV\in \mathcal{H}_{\text{row}}} \frac{\lambda}{\rho}\|\bV\|_{\text{row}, q} + \frac{1}{2}\left\|\bV -(\bD_{\text{row}}\bU^{(k+1)} + \rho^{-1}\bZ^{(k)}_{\text{row}}) \right\|_{\mathcal{H}_2}^2 = \prox_{\lambda/\rho \|\cdot\|_{\text{row}, q}}\left(\bD_{\text{row}}\bU^{(k+1)} + \rho^{-1}\bZ^{(k)}_{\text{row}}\right).\]
A similar analysis gives the $\bV_{\text{col}}$-update (\emph{i.e.}, the $\bW$-update above):
\[\bV^{(k+1)}_{\text{col}} = \argmin_{\bW \in \mathcal{H}_{\text{col}}} \frac{\lambda}{\rho}\|\bW\|_{\text{col}, q} + \frac{1}{2}\left\|\bW - (\bU^{(k+1)}\bD_{\text{col}} +\rho^{-1}\bZ^{(k)}_{\text{col}} \right)\|_{\mathcal{H}_3}^2 = \prox_{\lambda/\rho \|\cdot\|_{\text{col}, q}}\left(\bU^{(k+1)}\bD_{\text{col}} + \rho^{-1}\bZ^{(k)}_{\text{col}}\right).\]
Finally, the $\bZ$-update is given by: 
\begin{align*}
  \bZ^{(k+1)} &= \bZ^{(k)} + \rho \left(\mathfrak{L}_1\bU^{(k+1)} + \mathfrak{L}_2\bV^{(k+1)} + \mathfrak{L}_3\bW^{(k+1)} - \bb\right) \\
  \begin{pmatrix} \bZ^{(k+1)}_{\text{row}} \\ \bZ^{(k+1)}_{\text{col}} \end{pmatrix} &= \begin{pmatrix} \bZ^{(k)}_{\text{row}} \\ \bZ^{(k)}_{\text{col}} \end{pmatrix} + \rho\left(\begin{pmatrix} \bD_{\text{row}}\bU^{(k+1)} \\ \bU^{(k+1)}\bD_{\text{col}} \end{pmatrix} + \begin{pmatrix} -\bV^{(k+1)}_{\text{row}} \\ \bzero_{\mathcal{H}_3} \end{pmatrix} + \begin{pmatrix} \bzero_{\mathcal{H}_2} \\ -\bV^{(k+1)}_{\text{col}} \end{pmatrix} - \begin{pmatrix} \bzero_{\mathcal{H}_2} \\ \bzero_{\mathcal{H}_3} \end{pmatrix}\right) \\
  &= \begin{pmatrix} \bZ^{(k)}_{\text{row}} + \rho(\bD_{\text{row}}\bU^{(k+1)} - \bV^{(k+1)}_{\text{row}})\\ \bZ^{(k)}_{\text{col}} + \rho(\bU^{(k+1)}\bD_{\text{col}} - \bV^{(k+1)}_{\text{col}})\end{pmatrix}
\end{align*}

\end{document}